\documentclass[runningheads]{llncs}

\usepackage{accv}

\usepackage{accvabbrv}

\usepackage{graphicx}
\usepackage{booktabs}

\usepackage[accsupp]{axessibility}  

\usepackage{algorithm}
\usepackage{algpseudocode}
\usepackage{amsmath}
\usepackage{amssymb}

\usepackage[pagebackref,breaklinks,colorlinks,citecolor=accvblue]{hyperref}

\usepackage{orcidlink}

\begin{document}

\title{DiffGround: Semantic Difference-Guided Zero-Shot Video Question Grounding} 

\titlerunning{DiffGround}

\author{Cheng-Han Lee\inst{1} \and
Amir Salimi\inst{2} \and
Hsiang-Wei Huang\inst{2} \and
Che-Chun Su\inst{2} \and
Min Sun\inst{2} \and
Cheng-Hao Kuo\inst{2}
}

\authorrunning{Cheng-Han Lee et al.}

\institute{The University of Texas at Austin, United States
 \and
Amazon , United States}

\maketitle

Video Question Grounding (VideoQG) requires both accurate question answering and precise temporal localization. Existing supervised methods rely on costly temporal annotations, while existing zero-shot approaches are primarily designed for Video Question Answering (VQA) and are not optimized for temporal grounding. We propose DiffGround, an action-driven zero-shot framework built upon frozen Multimodal Large Language Models (MLLMs). Our framework first predicts an answer-relevant action cue that transforms the input question into an explicit semantic target for grounding. Action Cue converts a reasoning-oriented question into an action-centric retrieval query. Guided by the action cue, a Vision-Language Model (VLM) efficiently selects informative keyframes for reasoning. Finally, we introduce Semantic Diff Refinement, a training-free temporal refinement method that exploits semantic differences between neighboring SigLIP2 embeddings to recover complete action segments from sparse predictions. Extensive experiments on NExT-GQA \cite{xiao2024can} and STAR \cite{wu2024star} show that DiffGround consistently outperforms existing zero-shot and supervised methods, establishing a new state of the art for Video Question Grounding. We plan to publicly release the source code at  \href{https://github.com/steven413d/DiffGround}{\textcolor{blue}{https://github.com/steven413d/DiffGround}}.    
\section{Introduction}

\begin{figure}[t]
\begin{center}
 \includegraphics[width=0.9\linewidth]{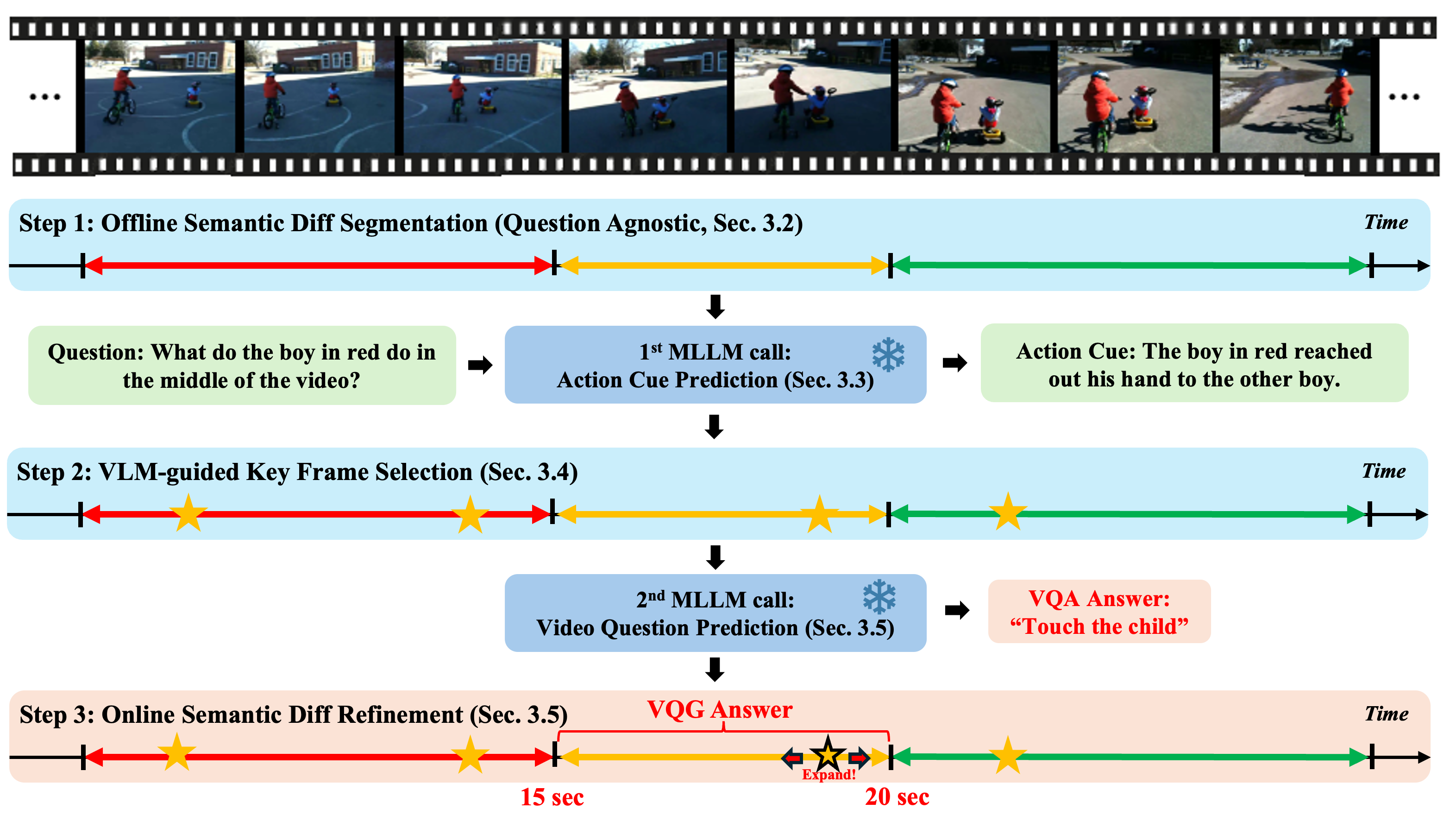}
\end{center}
\caption{\textbf{Overview of the proposed DiffGround framework.} Given a video and a question, DiffGround first performs offline, question-agnostic action segment preprocessing to partition the video into semantically coherent action segments. A frozen MLLM then predicts an action cue describing the target action with a single MLLM call. Guided by the question and predicted action cue, a Vision-Language Model (VLM) ranks video frames to select the most informative keyframes for Video Question Answering (VQA). Finally, Semantic Diff Refinement expands the initial grounding prediction by leveraging SigLIP2 semantic differences to align temporal boundaries with the complete action segment, producing more accurate Video Question Grounding (VideoQG) results.}
\label{fig:long}
\label{fig:onecol}
\label{fig1}
\end{figure}  

Video Question Grounding (VideoQG) is an emerging multimodal task that requires models to not only answer questions about video content but also accurately localize the temporal segments supporting their answers. Compared with conventional Video Question Answering (VQA), VideoQG requires both semantic reasoning and fine-grained temporal localization, making it a fundamental challenge for video understanding with applications in video retrieval, human-computer interaction, and embodied AI.

Existing approaches can be broadly categorized into supervised and zero-shot methods, both of which suffer from important limitations. Supervised methods rely on expensive temporal annotations and often exhibit a trade-off between VQA accuracy and grounding precision. More recently, Multimodal Large Language Models (MLLMs) have enabled zero-shot VideoQG without additional training. However, existing methods are primarily designed for VQA rather than temporal grounding, limiting their ability to accurately localize actions.

We identify two fundamental challenges underlying existing zero-shot approaches. First, they directly use the input question as the retrieval query for frame selection. However, natural language questions are designed for reasoning rather than visual retrieval, containing interrogative expressions, temporal references, and contextual descriptions that provide limited guidance for identifying relevant visual evidence. As a result, current methods tend to retrieve frames containing the correct objects instead of the target action, leading to object-centric rather than action-centric frame selection \cite{tstar}. Second, sparse keyframe predictions alone are insufficient for accurate temporal localization. Existing approaches often compensate with dense frame sampling or multiple MLLM calls, resulting in substantial computational cost while still failing to recover complete action boundaries.

To address these challenges, we propose DiffGround, an action-driven zero-shot Video Question Grounding framework, illustrated in Fig.~\ref{fig1}. DiffGround first performs an offline action segmentation by measuring semantic differences between neighboring Vision-Language Model (VLM) embeddings, producing reusable action segments for each video. Given a question, a frozen MLLM performs an initial reasoning step to generate an Action Cue, a concise action-centric description that serves as an explicit retrieval target. Guided by both the question and the predicted action cue, the VLM efficiently identifies informative keyframes. Finally, Semantic Diff Refinement expands sparse grounding predictions using the precomputed semantic transitions to recover complete action boundaries.

Our framework is built upon three key innovations. First, we introduce an Action Cue Prediction module that converts reasoning-oriented questions into concise action-centric representations, substantially improving frame retrieval quality with only a single MLLM inference. Second, we propose a VLM-guided keyframe selection strategy that jointly matches the question and predicted action cue, enabling efficient action-aware frame retrieval. Third, we introduce Semantic Diff Refinement, a training-free temporal refinement method that exploits semantic differences between adjacent SigLIP2 embeddings to recover complete action segments. Compared with conventional motion-based segmentation methods, Semantic Diff directly captures semantic transitions, operates in the semantic embedding space rather than relying on low-level motion cues, and requires minimal hyperparameter tuning. Since visual embeddings are extracted once and reused across multiple questions associated with the same video, the preprocessing is performed entirely offline, significantly reducing online inference cost.

Extensive experiments on NExT-GQA \cite{xiao2024can} and STAR \cite{wu2024star} demonstrate that DiffGround consistently outperforms existing zero-shot and supervised approaches, establishing a new state of the art in both Video Question Grounding and Video Question Answering.
\section{Related Works}
\subsection{Caption and Agent based Methods}
Recent zero-shot frameworks leverage textual intermediate representations to facilitate long-video understanding. LLoVi~\cite{zhang2023simple} and MoReVQA~\cite{min2024morevqa} first generate dense video captions and index them into textual memories for retrieval-based reasoning. Meanwhile, agent-based frameworks, such as VideoAgent~\cite{wang2024videoagent} and BOLT~\cite{liu2025bolt}, formulate video understanding as an iterative planning process, where an MLLM repeatedly queries localized video segments through multiple reasoning steps.

These approaches demonstrate strong zero-shot reasoning capability by exploiting the language understanding ability of MLLMs. However, they generally require numerous MLLM calls during either offline indexing or online planning, leading to considerable computational overhead. Moreover, because temporal reasoning is performed primarily through textual representations, the resulting temporal localization is constrained by the granularity of generated captions, making precise boundary localization challenging. In contrast, our approach directly operates on visual representations while requiring only a single MLLM inference for each query.

\subsection{Training-based Methods}

Training-based Video Question Grounding methods can be broadly categorized into weakly supervised and temporally supervised approaches.

Weakly supervised methods, such as CRA~\cite{chen2025cross} and TOGA~\cite{gupta2025toga}, are trained solely with video-level question-answer pairs. While avoiding dense temporal annotations, they typically rely on coarse visual-textual correlations and often struggle to precisely localize action boundaries.

Temporally supervised approaches explicitly exploit temporal annotations to improve grounding performance. Representative methods include ActPrompt~\cite{wang2026actprompt}, which introduces action-centric prompts to guide temporal attention, and fully supervised frameworks such as Grounded-VideoLLM~\cite{wang2024grounded}, VideoStreaming~\cite{qian2024streaming}, and SeViLA~\cite{yu2023self}, which fine-tune large video-language models using dense temporal annotations.

Unlike these methods, DiffGround is a training-free framework that requires no parameter updates or temporal annotations. Instead, it performs action-aware grounding through Action Cue Prediction and Semantic Diff Refinement, enabling strong generalization to reasoning-oriented VideoQG tasks.

\subsection{Frame Sampling-based Methods}
To reduce the computational cost of long-video understanding, recent zero-shot methods perform sparse frame sampling before MLLM inference. Representative approaches include TStar~\cite{tstar}, Adaptive Key Frame Sampling~\cite{tang2025adaptive}, and Q-Frame~\cite{zhang2025q}, which select informative frames based on spatial saliency, cross-attention, or vision-language similarity. By avoiding dense video rendering and task-specific fine-tuning, these methods provide an efficient solution for zero-shot VideoQA.

However, existing frame sampling strategies are primarily designed for question answering rather than temporal grounding. First, they retrieve discrete keyframes without explicitly recovering continuous temporal spans. Consequently, temporal boundaries must be inferred from sparse observations, often leading to inaccurate localization. Second, their retrieval strategies remain largely object-centric, emphasizing visual entities instead of action dynamics. As a result, they may successfully identify relevant objects while missing the temporal evolution of the target action. Our proposed DiffGround addresses both challenges through Action Cue Prediction for action-aware frame retrieval and Semantic Diff Refinement for recovering complete action segments from sparse predictions.

\begin{figure*}[t]
\begin{center}
 \includegraphics[width=1.0\linewidth]{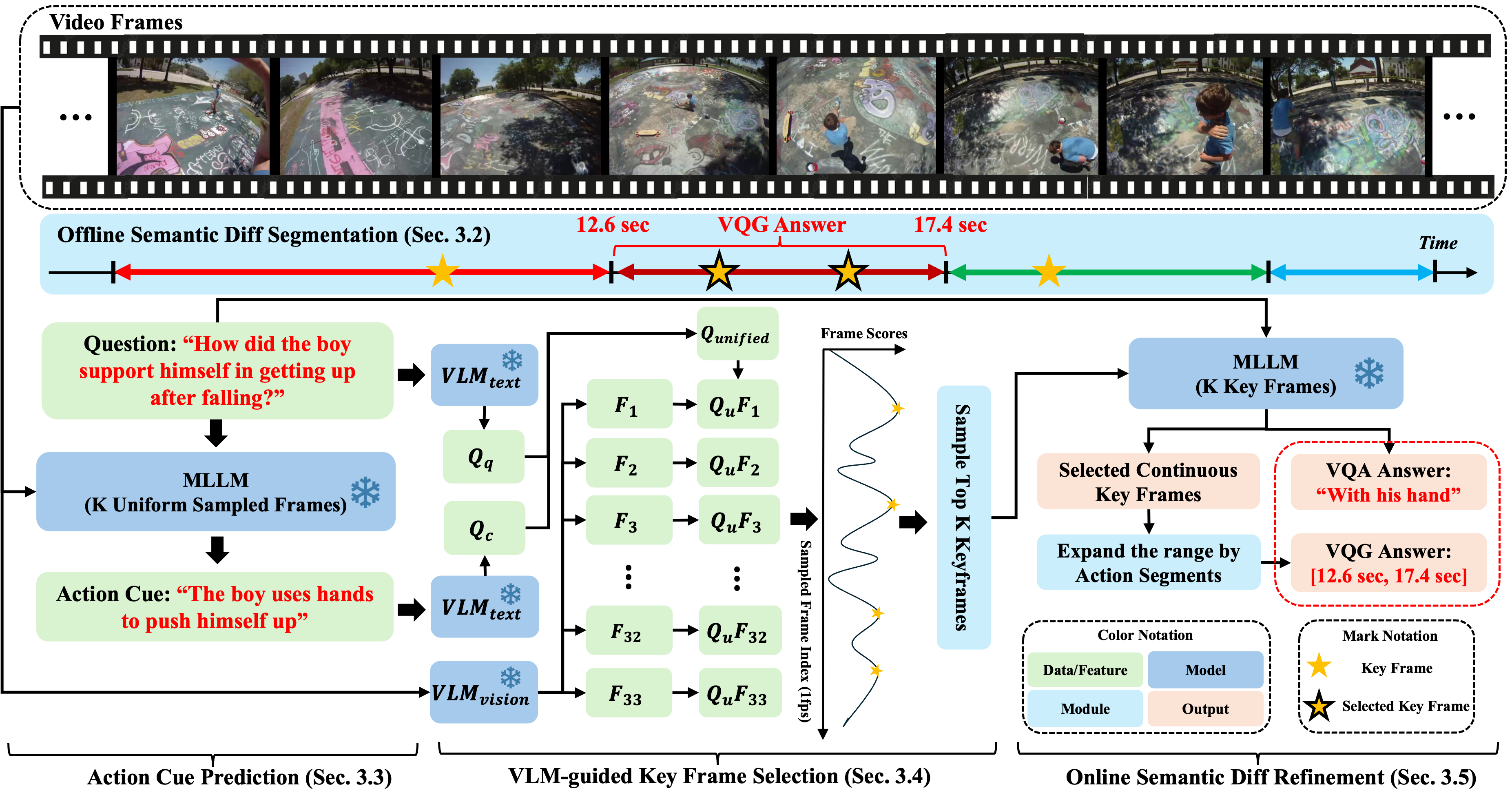}
\end{center}
\caption{\textbf{Overview of the proposed DiffGround framework.}
DiffGround consists of three stages. First, a frozen MLLM predicts an \emph{Action Cue} from the input question and uniformly sampled video frames, converting a reasoning-oriented question into an action-centric retrieval query. Second, a frozen VLM (SigLIP2) jointly matches the question and Action Cue to retrieve the Top-$K$ most informative keyframes. Meanwhile, an offline Semantic Diff Segmentation module partitions the video into semantically coherent action segments. Finally, the retrieved keyframes are used for Video Question Answering, and the predicted timestamps are refined by the precomputed semantic segments to recover complete temporal boundaries for Video Question Grounding.}
\label{fig:long}
\label{fig:onecol}
\label{fig2}
\end{figure*}  
\section{Method}
\subsection{Overview}

Given a video $\mathcal{V}$ and a question $q$, DiffGround predicts both the answer and its supporting temporal segment through an offline preprocessing stage and an online grounding stage, as illustrated in Fig.~\ref{fig2}.

During offline preprocessing, we partition the video into semantically coherent action segments by measuring semantic differences between neighboring SigLIP2 \cite{tschannen2025siglip} visual embeddings. These action segments are computed once for each video and reused for all subsequent questions.

During online inference, a frozen MLLM first performs an initial reasoning step to generate an \textit{Action Cue}, a concise action-centric description of the target event. The Action Cue is combined with the original question to retrieve informative keyframes using a Vision-Language Model (VLM). The selected keyframes are then fed into the MLLM to predict the final answer. Finally, Semantic Diff Refinement aligns the retrieved keyframes with the precomputed action segments to recover accurate temporal boundaries.

\subsection{Offline Semantic Diff Segmentation}
The Top-$K$ keyframes selected by the VLM provide sparse observations of the target action but do not directly reveal its complete temporal extent. We therefore seek to partition each video into semantically coherent action segments, allowing sparse frame predictions to be naturally expanded into complete temporal intervals during inference.

Instead, we propose \textbf{Semantic Diff Refinement}, a training-free temporal refinement framework that leverages semantic transitions in a Vision-Language Model (VLM) embedding space. The key observation is that modern VLMs encode high-level semantic information, allowing changes between neighboring embeddings to naturally indicate scene transitions and action boundaries. Since the visual embeddings are computed once for each video, the entire refinement process is performed offline and reused for all subsequent questions.

\subsubsection{Semantic Diff Feature Extraction}

Given an input video $\mathcal{V}$, we first partition it into $N$ consecutive one-second clips,

\begin{equation}
\mathcal{C}=\{c_1,c_2,\ldots,c_N\}.
\end{equation}

For each clip $c_i$, we uniformly sample $k$ frames,

\[
\mathcal{F}_i=\{f_{i,1},\ldots,f_{i,k}\},
\]

and extract their visual embeddings using the frozen SigLIP2 vision encoder,

\begin{equation}
v_{i,j}=f_{\mathrm{VLM}}(f_{i,j})\in\mathbb{R}^{d}.
\end{equation}

Instead of explicitly estimating motion, we characterize temporal changes directly in the VLM embedding space. For the $i$-th clip, we define its semantic displacement as

\begin{equation}
s_i =
\frac{v_{i,k}-v_{i,1}}{k-1},
\end{equation}

where $v_{i,1}$ and $v_{i,k}$ denote the embeddings of the first and last sampled frames, respectively. The resulting sequence

\[
\mathcal{S}=\{s_1,s_2,\ldots,s_N\}
\]

provides a compact representation of semantic changes across consecutive video clips.

\subsubsection{Adaptive Greedy Semantic Segmentation}

Given the sequence $\mathcal{S}$, our objective is to partition the video into semantically coherent action segments.

Rather than specifying the number of segments manually, we first estimate the target number of segments using semantic change-point detection. For each pair of adjacent clip representations, we compute

\begin{equation}
d_i=\|s_{i+1}-s_i\|_2.
\end{equation}

Large values of $d_i$ indicate significant semantic transitions. We therefore estimate an adaptive threshold,

\begin{equation}
\tau=\mu_D+\alpha\sigma_D,
\end{equation}

where $\mu_D$ and $\sigma_D$ denote the mean and standard deviation of $\{d_i\}$, respectively. The target number of action segments is then estimated as

\begin{equation}
M=
|\{d_i>\tau\}|+1.
\end{equation}

Starting from $N$ atomic clips, we greedily merge neighboring segments with the highest semantic similarity. Specifically, the adjacent pair

\begin{equation}
i^*=
\arg\max_i
\cos(s_i,s_{i+1})
\end{equation}

is merged during each iteration. The feature of the merged segment is updated using a duration-weighted average to preserve semantic consistency. The merging process continues until the number of remaining segments equals the estimated target value $M$.

\begin{algorithm}[t]
\footnotesize
\caption{Adaptive Greedy Semantic Diff Segmentation}
\label{alg:greedy_merge}
\begin{algorithmic}[1]

\Require
Semantic clip features
$\mathcal{S}=\{s_1,\ldots,s_N\}$,
threshold coefficient $\alpha$

\Ensure
Semantic action segments $\mathcal{A}$

\Statex
\textbf{Stage I: Estimate Target Segment Number}

\For{$i=1$ to $N-1$}
    \State $d_i \leftarrow \|s_{i+1}-s_i\|_2$
\EndFor

\State $\tau \leftarrow \mu_D+\alpha\sigma_D$

\State $M \leftarrow |\{d_i>\tau\}|+1$

\Statex

\textbf{Stage II: Greedy Semantic Merging}

\State Initialize
$\mathcal{A}=\{(c_i,s_i)\}_{i=1}^{N}$

\While{$|\mathcal{A}|>M$}

    \State
    $i^*\leftarrow
    \arg\max_i
    \cos(\mathcal{A}[i].f,
         \mathcal{A}[i+1].f)$

    \State Merge
    $\mathcal{A}[i^*]$
    and
    $\mathcal{A}[i^*+1]$

    \State Update feature using duration-weighted average

    \State Remove
    $\mathcal{A}[i^*+1]$

\EndWhile

\State \Return $\mathcal{A}$

\end{algorithmic}
\end{algorithm}

Algorithm~\ref{alg:greedy_merge} summarizes the complete adaptive greedy segmentation procedure.

\subsection{Action Cue Prediction}

Existing zero-shot methods directly use the input question as the retrieval query for frame selection. However, natural language questions are designed for reasoning rather than visual retrieval, often containing interrogative phrases, temporal references, and contextual descriptions that dilute the retrieval signal. Consequently, the retrieved frames frequently correspond to relevant objects instead of the target action.

To address this issue, we introduce an \textbf{Action Cue Prediction} module that performs an initial reasoning step before frame retrieval. Given the input question $q$ and a small set of uniformly sampled frames
$\mathcal{F}_{\mathrm{uniform}}$, a frozen MLLM generates a concise action-centric description:

\begin{equation}
c=f_{\mathrm{MLLM}}(q,\mathcal{F}_{\mathrm{uniform}}),
\end{equation}

where $c$ denotes the predicted Action Cue.

Unlike the original question, which is optimized for reasoning, the Action Cue explicitly describes the target action (e.g., ``The boy in red touches the child''), providing a semantically focused retrieval target for subsequent frame selection.

\subsection{VLM-guided Keyframe Selection}

Given a set of candidate frames
$\mathcal{F}=\{f_i\}_{i=1}^{T}$,
we encode each frame using the SigLIP2 vision encoder,

\begin{equation}
F=\mathrm{VLM}_{vision}(\mathcal{F})\in\mathbb{R}^{T\times d}.
\end{equation}

The question and Action Cue are independently encoded using the SigLIP2 text encoder,

\begin{align}
Q_q &= \mathrm{VLM}_{text}(q),\\
Q_c &= \mathrm{VLM}_{text}(c),
\end{align}

and combined into a unified representation,

\begin{equation}
Q=Q_q+Q_c.
\end{equation}

Frame relevance is computed using cosine similarity,

\begin{equation}
S=QF^T,
\end{equation}

and the Top-$K$ highest-scoring frames are selected for downstream VQA.

The Action Cue complements the original question by emphasizing action semantics, while the question preserves contextual information. Their combination provides a stronger retrieval signal than either representation alone.

\subsection{Online Semantic Diff Refinement}

The semantic segmentation described above is performed entirely offline and shared by all questions associated with the same video.


During online inference, the VLM first retrieves the Top-$K$ keyframes using the unified query representation described in Section~3.4. As illustrated in Fig.~\ref{fig2}, the selected keyframes are then fed into a second MLLM call to jointly predict the answer and the supporting keyframe indices. Each predicted keyframe is subsequently mapped to its corresponding precomputed semantic action segment. Instead of returning isolated frame timestamps, DiffGround expands the sparse keyframe predictions to the complete action segments, producing temporally coherent grounding results with negligible additional computation.

\begin{figure*}[t]
\begin{center}
\includegraphics[width=1.0\linewidth]{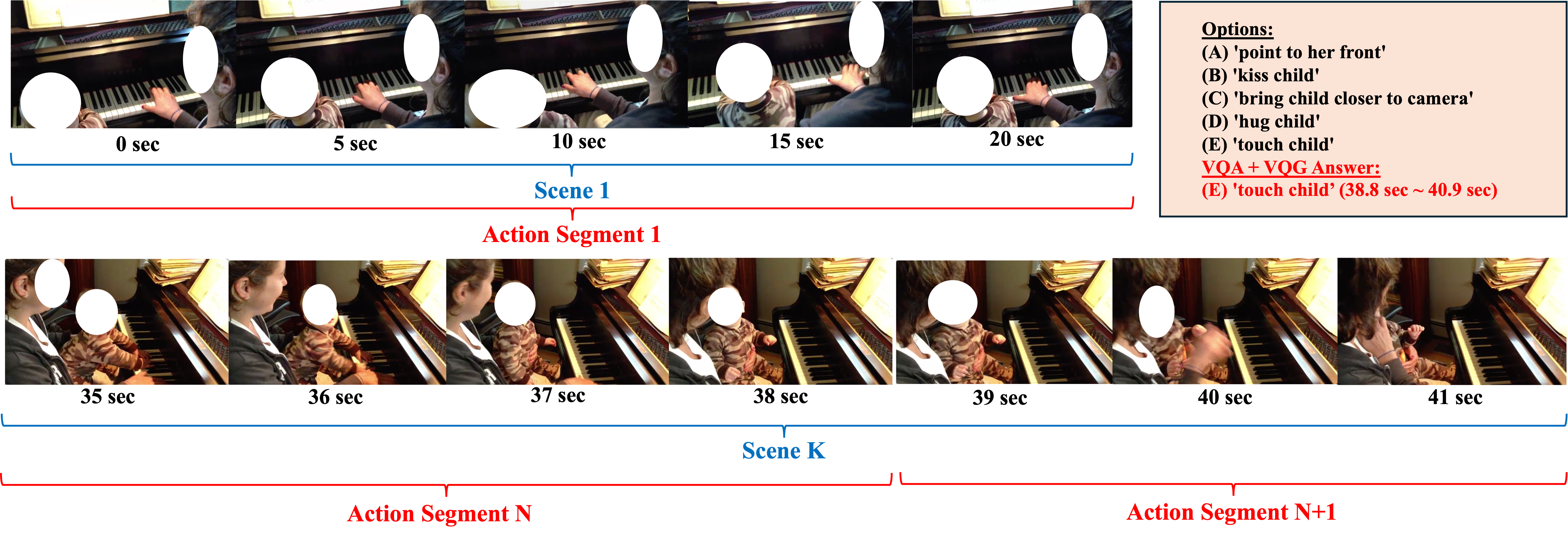}
\end{center}
\caption{\textbf{Visualization of the proposed Semantic Diff segmentation.}
By measuring semantic transitions between neighboring clips, our offline preprocessing partitions the video into semantically coherent action segments.
Besides separating distinct scenes (e.g., piano playing versus woman and child), it further identifies fine-grained action boundaries, isolating the brief ``touch child'' interaction (38.8--40.9 s) from surrounding activities.
During online inference, Semantic Diff Refinement aligns the retrieved keyframe with the corresponding action segment, producing accurate temporal grounding.}
\label{fig3}
\end{figure*}

Figure~\ref{fig3} illustrates a representative example. Besides separating two distinct scenes (``playing piano'' and ``woman with child''), Semantic Diff further identifies the brief ``touch child'' interaction as an independent action segment. Although the retrieved keyframe corresponds to only a single instant, Semantic Diff Refinement successfully recovers the entire temporal interval (38.8--40.9 s), enabling accurate Video Question Grounding.

To the best of our knowledge, DiffGround is the first zero-shot Video Question Grounding framework to leverage semantic differences in Vision-Language Model (VLM) embeddings for action segmentation. Since semantic features are precomputed offline, Semantic Diff efficiently captures both scene transitions and fine-grained action boundaries with negligible online overhead.

\section{Experiments}

\subsection{Datasets}

We evaluate our method on two standard benchmarks: the NExT-GQA test set \cite{xiao2024can} and the STAR validation set \cite{wu2024star}, as its test set is not publicly released. Detailed statistics for both datasets are presented in Table~\ref{tab1}.

\noindent\textbf{NExT-GQA \cite{xiao2024can}.}
NExT-GQA selects long videos with an average length of 40 seconds. Each video has multiple questions, and the answers’ grounding can overlap. The videos include a sequence of atomic events, and the questions involve interaction between multiple actors and objects. Questions are of two types: causal why/how questions and temporal questions with when/before/after clauses. Causal questions, such as Why are there two men standing at the center island and holding their camera?, require localizing the evidence of the answer, which is Recording for this question. Temporal questions, such as What did the boy do after the green man walked past him?, require understanding the temporal evolution of the events to generate an answer such as Look at man in green. The train set consists of 3,870 videos and 34,132 QA pairs without the grounding annotations. The test set consists of 990 videos with 5,553 QA pairs. Though the dataset provides multiple-choice answers, we generate open-ended responses without observing the options.

\noindent\textbf{STAR \cite{wu2024star}.}
STAR consists of about 60K situated reasoning questions with programs and answers, 240K candidate choices, and 22K trimmed situation video clips. Situation video clips in the benchmark are sourced from human activity videos, which record the dynamic interaction processes of human actions and surrounding environments in daily-life scenes. It also provides about 144K situation hypergraphs as structured situation abstraction. Designed questions cover four types of skills for situated reasoning. The dataset includes annotated questions with answers and options. Each question answering corresponds to a specific program for reasoning logic. To connect situation abstraction and reasoning diagnosis for question-answering, STAR provides situation hypergraphs tied with executable programs. Then situations, questions, and options are aligned with the unified data type schema, including actions, objects, humans, and relations. The STAR includes 111 action predicates, 28 objects, and 24 relationships. The benchmark is split into training/validation/test sets with a ratio of about 6:1:1.

\begin{table}[t]
\centering
\caption{\textbf{Statistic of the evaluation sets.} Note: We utilize NExT-GQA \cite{xiao2024can}'s test set and STAR \cite{wu2024star}'s validation set for evaluation. By employing a column-width resizing wrapper, this table maintains a consistent style and layout alignment with our main quantitative results.}
\label{tab1}
\setlength{\tabcolsep}{6.0pt} 
\resizebox{\columnwidth}{!}{
\begin{tabular}{l | c c c c c c} 
\hline
Dataset & Vid. & Que. & Seg. & Seg. Dur.(s) & Vid. Dur.(s) & Ratio (S./V.) \\ 
\hline
\hline
NExT-GQA & 990 & 5553 & 6600 & 6.7 & 39.5 & 0.2 \\
STAR & 914 & 7098 & 7098 & 11.9 & 30.0 & 0.4 \\
\hline
\end{tabular}
}
\end{table}

\subsection{Evaluation Metrics}
We consider five metrics: intersection over union (IoU), intersection over prediction (IoP), IoU@0.5, IoP@0.5, and Acc@GQA proposed in NExT-GQA \cite{xiao2024can}. IoU measures the overlap between ground truth and predicted groundings. IoP measures the portion of predicted grounding containing the ground truth, similar to precision. While these metrics focus on grounding, Acc@GQA focuses on both the correctness of the answers and correct grounding with IoP $\leqslant$ 0.5.  Apart from these metrics, we consider the Acc@QA metric for QA performance.

\subsection{Implementation Details}

\noindent\textbf{Experimental Setup.}
We evaluate DiffGround on the NExT-GQA test set (5,553 video-question pairs) and the STAR validation set (7,098 video-question pairs). Unless otherwise specified, GPT-4o~\cite{hurst2024gpt} is used as the Multimodal Large Language Model (MLLM), while SigLIP2 serves as the Vision-Language Model (VLM). The temperature is set to 0.0 for Action Cue Prediction to ensure deterministic outputs and 0.2 for the final Video Question Answering inference.

\noindent\textbf{Offline Semantic Segmentation.}
The only dataset-specific hyperparameter is the threshold coefficient $\alpha$ used in Algorithm~\ref{alg:greedy_merge}. We set $\alpha=-0.25$ for NExT-GQA and $\alpha=0.25$ for STAR according to the average ground-truth answer duration of each dataset. Smaller values produce finer-grained action segments, whereas larger values generate coarser temporal partitions.

\noindent\textbf{Implementation Details.}
All experiments are conducted on servers equipped with two NVIDIA RTX 4090 GPUs (24GB memory each). Additional implementation details, prompt templates, comparisons with alternative VLMs (CLIP and SigLIP2) and MLLMs (e.g., Qwen2.5-VL and Qwen3-VL), additional ablation studies, and the proposed \textbf{Acc@GQA (IoU)} evaluation metric are provided in the supplementary material.

\begin{figure}[t]
\begin{center}
 \includegraphics[width=1.0\linewidth]{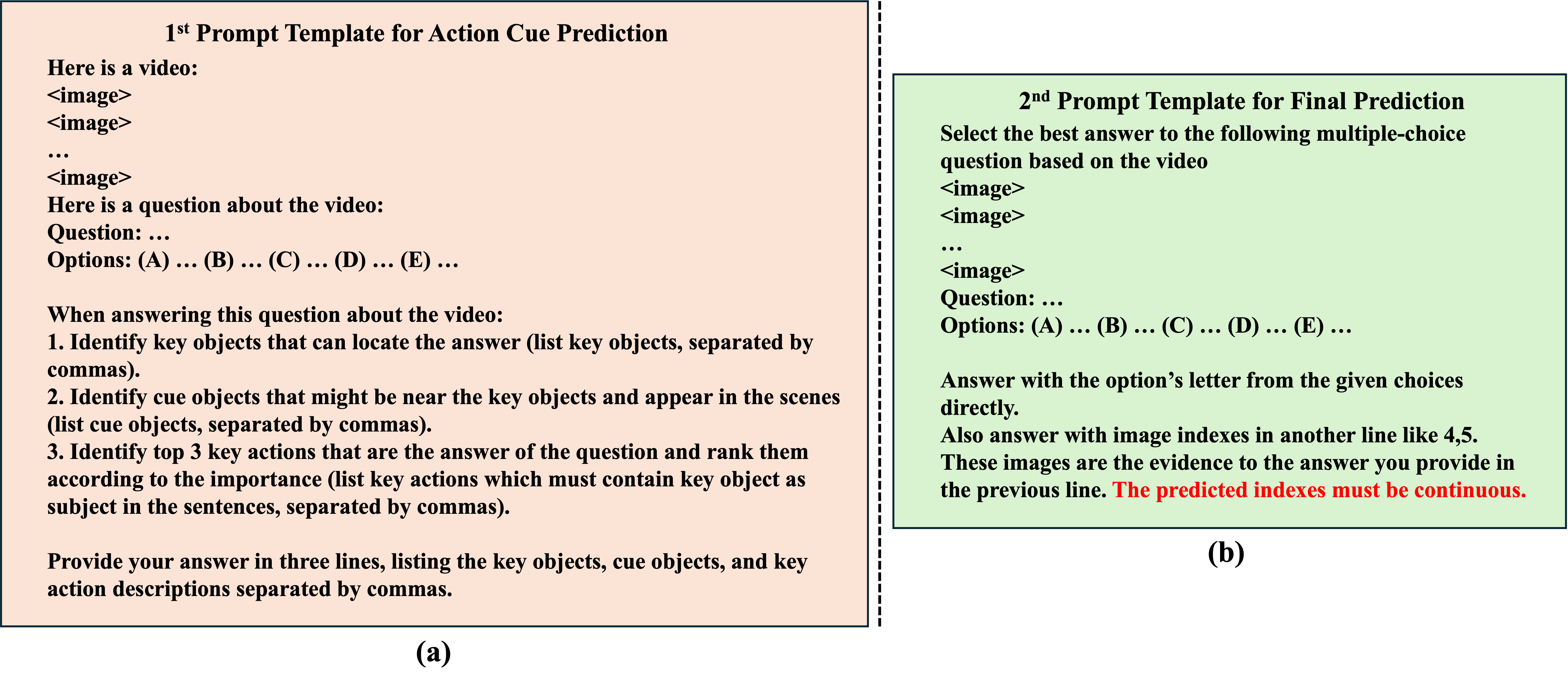}
\end{center}
\caption{\textbf{Prompt Template Design.} The prompt for our final VQG prediction, which is adapted from TStar \cite{tstar}.}
\label{fig:long}
\label{fig:onecol}
\label{fig4}
\end{figure}  

\subsection{Frame Retrieval Variants}

To analyze the contribution of different retrieval queries, we compare four frame retrieval variants while keeping the VLM retrieval pipeline unchanged.

\noindent\textbf{Uniform Sampling.}
Frames are sampled uniformly across the video and directly passed to the VQG prompt (Fig.~\ref{fig4}(b)) without Action Cue Prediction.

\noindent\textbf{Question Only.}
Frame relevance is computed using only the question embedding, $Q_q$. The selected frames are then processed using the VQG prompt (Fig.~\ref{fig4}(b)).

\noindent\textbf{Action Cue Only.}
The Action Cue is first generated using the Action Cue Prediction prompt (Fig.~\ref{fig4}(a)). Frame relevance is computed using only the Action Cue embedding, $Q_c$, and the selected frames are fed into the VQG prompt (Fig.~\ref{fig4}(b)).

\noindent\textbf{Object Cue Only.}
The Action Cue Prediction prompt (Fig.~\ref{fig4}(a)) is modified to output a single object noun (e.g., ``boy'') instead of an action description. Frame relevance is computed using the corresponding object embedding, and the selected frames are processed using the VQG prompt (Fig.~\ref{fig4}(b)).

Detailed prompt templates for all variants are provided in the supplementary material.

\begin{table}[t]
\centering
\caption{\textbf{Quantitative Results on NExT-GQA.} Note: $N$ means the number of sampled frames with 1 fps. * indicates that the methods do not require any post-training. By employing a column-width resizing wrapper, all model names are preserved in full within the single-column layout.}
\label{tab2}
\setlength{\tabcolsep}{3.0pt} 
\resizebox{\columnwidth}{!}{
\begin{tabular}{l | c | c | c | c c c | c} 
\hline
Method & Venue & Model & \# of frames & Acc@GQA ($\uparrow$) & mIoP ($\uparrow$) & mIoU ($\uparrow$) & \# of calls \\ 
\hline
\hline
IGV \cite{li2022invariant} & CVPR'22 & - & 16 & 10.2 & 21.4 & 14.0 & - \\
SeViLA \cite{yu2023self} & NIPS'23 & BLIP2 & \textbf{4} & 16.6 & 29.5 & 21.7 & - \\
*LLoVi \cite{zhang2023simple} & EMNLP'24 & GPT-4 + CogAgent  & - & 26.8 & 39.4 & 21.5 & $\approx N$ \\
VideoStreaming \cite{qian2024streaming} & NIPS'24 & Phi-2 2.7B + Vicuna-7B & 16 & 17.8 & 32.2 & 19.8 & 1 \\
CRA \cite{chen2025cross} & CVPR'25 & Temp[CLIP] & 32 & 18.2 & 28.6 & 14.2 & - \\
CRA \cite{chen2025cross} & CVPR'25 & FrozenBiLM & 32 & 18.8 & 26.5 & 13.5 & - \\
Grounded-VideoLLM \cite{wang2024grounded} & EMNLP'25 & Phi3.5 & 96 & 26.7 & 34.5 & 21.1 & 1 \\
TOGA \cite{gupta2025toga} & ICCV'25 & - & 16 & 24.6 & \underline{40.5} & \underline{24.6} & - \\
\hline
*Baseline (Uniform) & - & GPT-4o & \underline{8} & \underline{30.5} & 38.9 & \textbf{25.5} & 1 \\
*\textbf{Our} & - & GPT-4o + SigLIP2 & \underline{8} & \textbf{35.0} & \textbf{42.7} & \textbf{25.5} & 2 \\
\hline
Human & - & - & - & 82.1 & 72.1 & 61.2 & - \\
\hline
\end{tabular}
}
\end{table}

\begin{table}[t]
\centering
\caption{\textbf{Quantitative Results on STAR.} Note: * indicates that the methods do not require any post-training. }
\label{tab3}
\setlength{\tabcolsep}{3.0pt} 
\resizebox{\columnwidth}{!}{
\begin{tabular}{l | c | c | c | c c c c | c} 
\hline
Method & Venue & Model & \# of frames & Acc@GQA ($\uparrow$) & Acc@QA ($\uparrow$) & IoP@5 ($\uparrow$) & IoU@5 ($\uparrow$) & \# of calls \\ 
\hline
\hline
IGV \cite{li2022invariant} & CVPR'22 & - & \underline{16} & 13.1 & 31.7 & 42.8 & 1.7 & - \\
SeViLA \cite{yu2023self} & NIPS'23 & BLIP2 & \textbf{4} & 17.0 & 45.5 & 37.6 & \underline{19.5} & - \\
CRA \cite{chen2025cross}  & CVPR'25 & Temp[CLIP] & 32 & 26.8 & 58.6 & \underline{44.5} & 5.5 & - \\
CRA \cite{chen2025cross} & CVPR'25 & FrozenBiLM & 32 & \underline{27.5} & \underline{60.5} & 43.1 & 5.1 & - \\
\hline
*Baseline (Uniform) & - & GPT-4o & \textbf{4} & 25.7 & 59.3 & 41.4 & 18.9 & 1 \\
*\textbf{Our} & - & GPT-4o + SigLIP2 & \textbf{4} & \textbf{33.3} & \textbf{61.1} & \textbf{51.9} & \textbf{20.1} & 2 \\
\hline
\end{tabular}
}
\end{table}
\begin{table}[t]
\centering
\caption{\textbf{Ablation Study of Our Approach on NExT-GQA.} Note: Percentage gains indicate relative improvements over the Baseline (Uniform) configuration. SD=Semantic Diff Refinement.}
\label{tab4}
\setlength{\tabcolsep}{3.0pt} 
\resizebox{\columnwidth}{!}{
\begin{tabular}{l | c | c c c c | c} 
\hline
Method & \# of frames & Acc@GQA ($\uparrow$) & Acc@QA ($\uparrow$) & mIoP ($\uparrow$) & mIoU ($\uparrow$) & \# of calls \\ 
\hline
\hline
Baseline (Uniform) & 8 & 30.5 & 77.9 & 38.9 & 25.5 & 1 \\
\hline
Question (Q) & 8 & 37.3 & 77.0 & 45.0 & 20.0 & 1 \\
Object Cue (OC) & 8 & 34.0 & 77.3 & 41.4 & 18.7 & 2 \\
Action Cue (AC) & 8 & 38.1 & 78.2 & 46.1 & 20.0 & 2 \\
\hline
Our (Q + AC) & 8 & 38.5 (+8\%) & 78.7 (+0.8\%) & 46.1 (+7.2\%) & 19.9 (-5.6\%) & 2 \\
Our (Q + AC + SD) & 8 & 35.0 (+4.5\%) & 78.7 (+0.8\%) & 42.7 (+3.8\%) & 25.5 (+0\%) & 2 \\
\hline
\end{tabular}
}
\end{table}
\begin{table}[t]
\centering
\caption{\textbf{Ablation Study of Our Approach on STAR.} Percentage gains indicate relative improvements over the Baseline (Uniform) configuration.  SD=Semantic Diff Refinement.}
\label{tab5}
\setlength{\tabcolsep}{3.0pt} 
\resizebox{\columnwidth}{!}{
\begin{tabular}{l | c | c c c c | c} 
\hline
Method & \# of frames & Acc@GQA ($\uparrow$) & Acc@QA ($\uparrow$) & IoP@5 ($\uparrow$) & IoU@5 ($\uparrow$) & \# of calls \\ 
\hline
\hline
Baseline (Uniform) & 4 & 25.7 & 59.3 & 41.4 & 18.9 & 1 \\
\hline
Question (Q) & 4 & 31.0 & 55.3 & 53.1 & 8.2 & 1 \\
Object Cue (OC) & 4 & 30.7 & 60.7 & 47.2 & 7.4 & 2 \\
Action Cue (AC) & 4 & 33.1 & 61.3 & 49.8 & 7.5 & 2 \\
\hline
Our (Q + AC) & 4 & 33.6 (+7.9\%) & 61.1 (+1.8\%) & 52.2 (+11.8\%) & 7.7 (-11.2\%) & 2 \\
Our (Q + AC + SD) & 4 & 33.3 (+7.6\%) & 61.1 (+1.8\%) & 51.9 (+11.5\%) & 20.1 (+1.2\%) & 2 \\
\hline
\end{tabular}
}
\end{table}

\subsection{Comparing to State-of-the Arts}
\textbf{NExT-GQA.}
Table~\ref{tab2} compares DiffGround with existing supervised and zero-shot approaches on NExT-GQA. Despite requiring only two MLLM calls and no additional training, DiffGround achieves the best overall performance across all evaluation metrics. Compared with the strongest zero-shot baseline, our method improves Acc@GQA by 4.5 percentage points while simultaneously achieving the highest mIoP and mIoU. These results demonstrate that Action Cue Prediction and Semantic Diff Refinement effectively improve both question answering and temporal localization without increasing computational complexity.

\textbf{STAR.}
Table~\ref{tab3} further validates the generalization of DiffGround on the STAR benchmark. Our framework consistently achieves the best performance in both question answering and grounding while using only two MLLM calls. Compared with uniform sampling, DiffGround significantly improves Acc@GQA and IoP@5 while also producing the highest IoU@5 among all compared methods. These results suggest that the proposed action-driven framework generalizes well across datasets with different video durations and reasoning styles.

\subsection{Effectiveness of Action Cue Enhanced Key Frame Sampling}
\label{sec46}
Table~\ref{tab4} and Table~\ref{tab5} analyze different frame retrieval strategies. Using the question alone substantially improves Acc@GQA over uniform sampling, confirming that semantic retrieval is considerably more effective than fixed sampling. However, this improvement comes at the expense of temporal grounding, as IoU consistently decreases on both datasets. This observation indicates that questions are optimized for reasoning rather than visual retrieval, often identifying a few highly discriminative frames while overlooking the complete temporal extent of the target action.

We further compare our Action Cue with a simplified Object Cue that contains only the primary object noun. Although both cues improve retrieval over the baseline, the Action Cue consistently achieves higher Acc@GQA and IoP. This demonstrates that explicitly modeling actions and object interactions provides a more informative retrieval target than isolated object descriptions.

Finally, combining the original question with the predicted Action Cue achieves the strongest retrieval performance across both datasets. Nevertheless, despite producing the highest Acc@GQA, the retrieved keyframes still exhibit relatively low IoU, suggesting that frame retrieval alone is insufficient for accurate temporal localization. This motivates the Semantic Diff Refinement introduced in the next section.

\subsection{Effectiveness of Semantic Diff Refinement}
Table~\ref{tab4} and Table~\ref{tab5} evaluate the effectiveness of the proposed Semantic Diff Refinement. Comparing ``Our (Question + Action Cue)'' with the complete DiffGround framework shows that Semantic Diff consistently improves temporal grounding performance on both NExT-GQA and STAR. In particular, mIoU increases from 19.9\% to 25.5\% on NExT-GQA and from 7.7\% to 20.1\% on STAR, demonstrating that the proposed refinement effectively recovers more complete action boundaries from sparse keyframe predictions.

Interestingly, Semantic Diff Refinement slightly decreases Acc@GQA (IoP) and mIoP while substantially improving IoU. This behavior reflects the different optimization objectives of Video Question Answering and Video Question Grounding. Action Cue Prediction is designed to retrieve a small number of highly discriminative frames that maximize question answering accuracy. Such compact predictions often achieve high overlap precision but cover only part of the target action, leading to relatively low temporal IoU. In contrast, Semantic Diff Refinement expands these sparse predictions to semantically coherent action segments, producing predictions that more closely align with the complete temporal extent of the ground-truth action.

Consequently, the refined segments may include additional contextual frames that contribute little to frame-level retrieval metrics, resulting in a slight reduction in Acc@GQA (IoP) and mIoP. However, they provide substantially more accurate temporal localization, as reflected by the consistent improvements in mIoU. These results suggest that optimizing frame retrieval alone is insufficient for Video Question Grounding, and that explicit temporal refinement is necessary to recover complete action segments.

A complementary analysis using an \textbf{IoU-based variant of Acc@GQA} is provided in the supplementary material, further illustrating the trade-off between retrieval-oriented and localization-oriented evaluation.
\section{Conclusions}
We introduced DiffGround, a training-free, action-driven framework for zero-shot Video Question Grounding. Our key insight is that transforming reasoning-oriented questions into action-centric representations and refining sparse predictions using semantic transitions enables accurate and efficient temporal grounding. Extensive experiments on NExT-GQA and STAR demonstrate that DiffGround establishes a new state of the art while remaining computationally efficient.

%
%
\bibliographystyle{splncs04}
\bibliography{main}

\clearpage

\section{Appendix: Ablation Study of Different VLMs}
In Table \ref{tab6}, we analyze the impact of different Vision-Language Model (VLM) backbones on our framework. We compare the strongest variants from three prominent model families: CLIP (ViT-bigG-14) \cite{radford2021learning}, SigLIP (ViT-SO400M-14) \cite{tschannen2025siglip}, and SigLIP2 (ViT-SO400M-16) \cite{tschannen2025siglip}. The results demonstrate that SigLIP2 yields the most robust overall performance. Driven by its superior ImageNet accuracy (84.1\%), it achieves the highest scores across the majority of tasks, securing 35.9\% Acc@GQA and 42.7\% mIoP. While CLIP remains competitive in mIoU, SigLIP2's significant advantages in VQA accuracy and intersection-over-prediction make it the optimal choice for our unified query selection and semantic difference calculation.

\begin{table}[t]
\tiny
\centering
\caption{\footnotesize
\textbf{Comparison of different VLMs on NExT-GQA.}
``Train Data'' denotes the pre-training dataset, and ``IN-1K'' denotes ImageNet-1K top-1 accuracy.}
\label{tab6}
\resizebox{\columnwidth}{!}{
\begin{tabular}{c c c | c | c c c c}
\hline
VLM & Data & IN-1K & \# of frames & Acc@GQA & Acc@QA & mIoP & mIoU \\
\hline
\hline
CLIP \cite{radford2021learning} & LAION-2B & 80.1 & 8 & \underline{34.2} & \underline{78.1} & 41.5 & \textbf{25.7} \\
SigLIP \cite{zhai2023sigmoid} & WebLI & \underline{83.1} & 8 & 33.9 & 77.8 & \underline{42.0} & 25.4 \\
SigLIP2 \cite{tschannen2025siglip} & WebLI & \textbf{84.1} & 8 & \textbf{35.9} & \textbf{78.7} & \textbf{42.7} & 25.5 \\
\hline
\end{tabular}
}
\end{table}

\section{Appendix: A New Evaluation Metric for Video Question Grounding: Acc@GQA (IoU)}
To provide a more rigorous evaluation of temporal precision, we introduce a complementary metric, Acc@GQA (IoU). The standard Acc@GQA metric \cite{xiao2024can} relies on Intersection over Prediction (IoP@5). While useful, this metric can be misleading: because it measures only precision (containment), an extremely short prediction—even a single timestamp—can achieve a perfect score as long as it falls within the ground truth. This fails to penalize models that capture only a fraction of the full event duration.

To measure how accurately the predicted segment aligns with the ground truth, we define Acc@GQA (IoU) using an Intersection over Union (IoU) threshold of 0.3. As shown in Table \ref{tab7}, this metric reveals the true impact of our pipeline. While adding the Semantic Difference Refinement (SD) causes a slight drop in the containment-focused Acc@GQA (IoP) (from 38.5\% to 35.0\%), it drives a substantial improvement in the alignment-focused Acc@GQA (IoU), boosting it from 24.5\% to 32.0\%. This confirms that our refinement module effectively expands the prediction to match the true action boundaries, trading "safe," partial containment for precise, full-event grounding.

\begin{table}[t]
\tiny
\centering
\caption{\footnotesize
\textbf{Complementary evaluation using Acc@GQA (IoU).}
SD=Semantic Diff Refinement.}
\label{tab7}
\resizebox{\columnwidth}{!}{
\begin{tabular}{c | c | c c c c c}
\hline
Method & \# of frames & Acc@GQA (IoP) & Acc@GQA (IoU) & Acc@QA & mIoP & mIoU \\
\hline
\hline
Q+AC & 8 & \textbf{38.5} & 24.5 & \textbf{78.7} & \textbf{46.1} & 19.9 \\
Q+AC+SD & 8 & 35.0 & \textbf{32.0} & \textbf{78.7} & 42.7 & \textbf{25.5} \\
\hline
\end{tabular}
}
\end{table}


\section{Appendix: Performance of Utilizing Open-source MLLMs}
In Table \ref{tab8}, we evaluate the generalization capability of our framework by replacing GPT-4o with state-of-the-art open-source MLLMs: Qwen2.5-VL-Instruct-7B \cite{bai2025qwen2} and Qwen3VL-Instruct-8B \cite{yang2025qwen3}. While these models achieve reasonable VQA performance (e.g., 74.8\% Acc@QA for Qwen2.5), they exhibit a significant drop in temporal grounding accuracy, with mIoU scores falling below 16\%. Qualitative analysis reveals that unlike GPT-4o \cite{hurst2024gpt}, these smaller open-source models struggle to follow the strict formatting constraints of our prompt. They frequently hallucinate repetitive or generic index ranges (e.g., consistently outputting "1, 2" or "4, 5") regardless of the video content, likely due to a lack of exposure to this specific grounding-via-indexing task during their instruction tuning. This highlights that while open-source MLLMs are closing the gap in visual reasoning, they still lag behind proprietary models in complex instruction following and structured output generation.


\begin{table}[t]
\tiny
\centering
\caption{\footnotesize
\textbf{Comparison of different MLLMs on NExT-GQA.}
All methods use the same retrieval pipeline; only the MLLM is changed.}
\label{tab8}
\resizebox{\columnwidth}{!}{
\begin{tabular}{c | c | c c c c}
\hline
MLLM & \# of frames & Acc@GQA & Acc@QA & mIoP & mIoU \\
\hline
\hline
Qwen2.5-VL \cite{bai2025qwen2} & 8 & 23.7 & 74.8 & 31.7 & 14.8 \\
Qwen3-VL \cite{yang2025qwen3} & 8 & 19.9 & 72.9 & 28.0 & 15.5 \\
\hline
GPT-4o & 8 & \textbf{35.0} & \textbf{78.7} & \textbf{42.7} & \textbf{25.5} \\
\hline
\end{tabular}
}
\end{table}

\section{Appendix: Prompt Template Design for Object Cue Predictor}
\begin{figure}[t]
\begin{center}
 \includegraphics[width=0.8\linewidth]{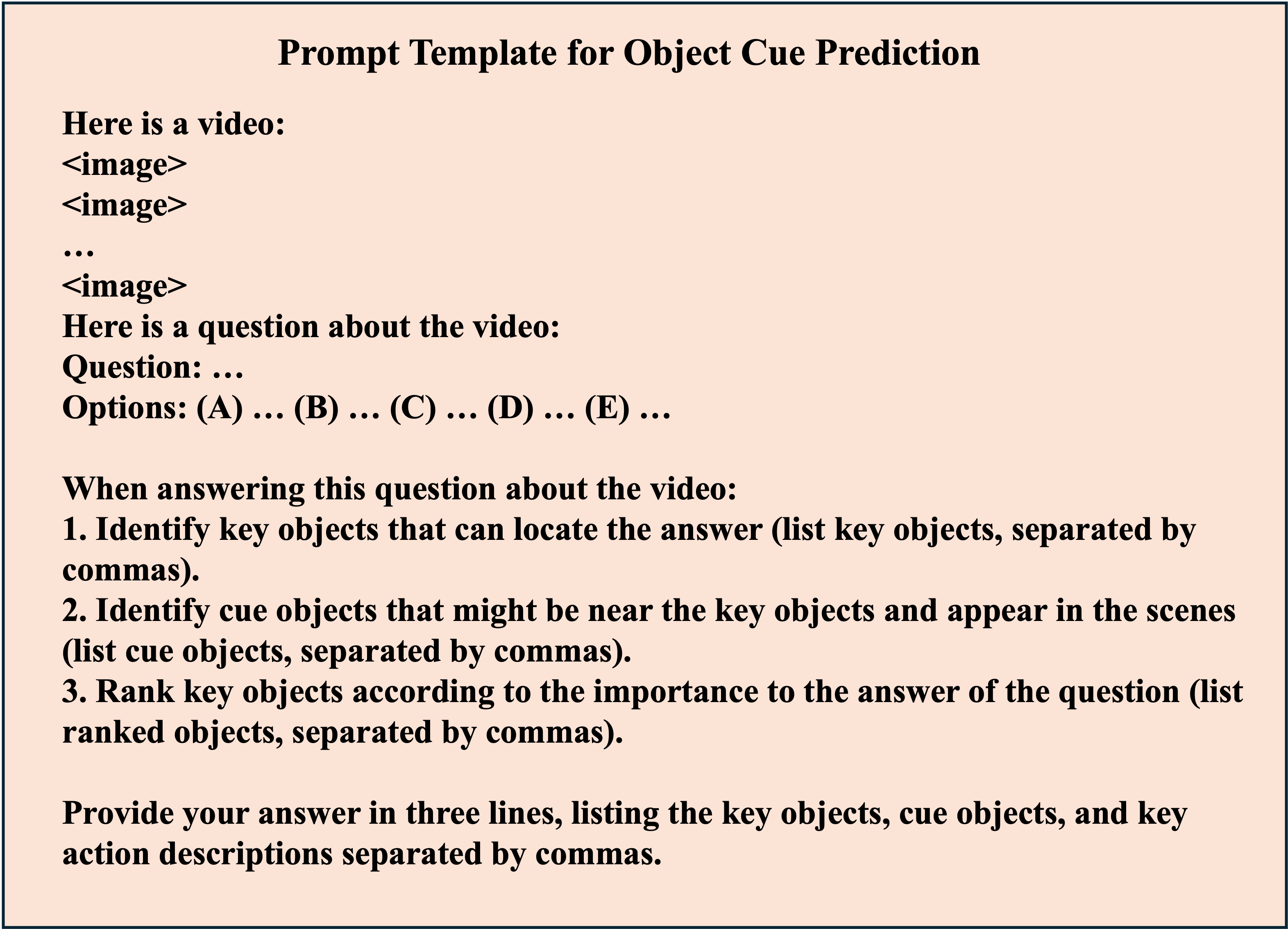}
\end{center}
\caption{\textbf{Prompt Template Design for Object Cue.} The prompt for generating object cues, which is adapted from TStar \cite{tstar}.}
\label{fig:long}
\label{fig:onecol}
\label{fig5}
\end{figure}

Figures \ref{fig5} illustrate the prompt templates designed for our Object Cue Predictor. For a detailed quantitative analysis of these components, please refer to the ablation study in Sec. 4.6.


\section{Appendix: Computational Cost}
Each MLLM call takes approximately $3$ secs. DiffGround requires two online MLLM calls plus approximately $1$ sec for retrieval/refinement, resulting in about 7 secs/query online latency. Offline SigLIP2 features and Semantic Diff segments are computed once per video and reused across questions.

\end{document}